\documentclass[a4paper,twoside]{article}

\usepackage{epsfig}
\usepackage{subcaption}
\usepackage{calc}
\usepackage{amssymb}
\usepackage{amstext}
\usepackage{amsmath}
\usepackage{amsthm}
\usepackage{multicol}
\usepackage{pslatex}
\usepackage{apalike}
\usepackage{SCITEPRESS}
\usepackage{multirow}
\usepackage{graphicx}
\usepackage{comment}
\usepackage[breaklinks]{hyperref} 
\usepackage{xcolor}
\begin{document}

\title{The MIS Check-Dam Dataset for Object Detection and Instance Segmentation Tasks}

\author{\authorname{Chintan Tundia\sup{1}\orcidAuthor{0000-0003-3169-1775}, Rajiv Kumar\sup{1}\orcidAuthor{0000-0003-4174-8587}, Om Damani \sup{1}\orcidAuthor{0000-0002-4043-9806} and G. Sivakumar \sup{1}\orcidAuthor{0000-0003-2890-6421}}
\affiliation{\sup{1}Indian Institute of Technology Bombay, Mumbai, INDIA}
\email{\{chintan, rajiv, damani\}@cse.iitb.ac.in, siva@iitb.ac.in}
}

\keywords{Object detection, Instance Segmentation, Remote Sensing, Image Transformers.}

\abstract{Deep learning has led to many recent advances in object detection and instance segmentation, among other computer vision tasks. These advancements have led to wide application of deep learning based methods and related methodologies in object detection tasks for satellite imagery. In this paper, we introduce MIS Check-Dam, a new dataset of check-dams from satellite imagery for building an automated system for the detection and mapping of check-dams, focusing on the importance of irrigation structures used for agriculture. We review some of the most recent object detection and instance segmentation methods and assess their performance on our new dataset. We evaluate several single stage, two-stage and attention based methods under various network configurations and backbone architectures. The dataset and the pre-trained models are available at 
\textit{\href{https://www.cse.iitb.ac.in/gramdrishti/about/}{https://www.cse.iitb.ac.in/gramdrishti/}}
}
\onecolumn \maketitle \normalsize \setcounter{footnote}{0} \vfill
\section{\uppercase{Introduction}}
\label{sec:introduction}
Machine learning and deep learning have evolved exponentially over the past decade contributing a lot to the fields of image and video processing~\cite{image_processing}, computer vision~\cite{visapp21_rajiv}, natural language processing~\cite{fedus2021switch}, reinforcement learning~\cite{henderson2018deep} and more. Understanding a scene involves detecting and identifying objects (object detection) or keypoints and different regions in images (instance and semantic segmentation). With the improvements in imaging technologies of the satellite sensors, large volumes of hyper-spectral and spatial resolution data are available, resulting in the application of object detection in remote sensing for identifying objects from satellite, aerial and SAR imagery. The availability of very large remote sensing datasets~\cite{diordataset} have led to applications like land monitoring, target identification, change detection, building detection, road detection, vehicle detection, and so on. 

\footnote{Authors \textit{a,b} made equal contribution }
With the increased spatial resolution, more objects are visible in satellite images, while been subjected to viewpoint variation, occlusion, background clutter, illumination, shadow, etc. This leads to various challenges in addition to the common object detection challenges like small spatial extent of foreground target objects, large scale search space, variety of perspectives and viewpoints, etc. The objects in remote sensing images are also subject to rotation, intra-class variations, similarity to surroundings, etc. making object detection on remote sensing images a challenging task. 
\begin{figure}[h]
     \begin{center}
     \includegraphics[width=0.98\linewidth]{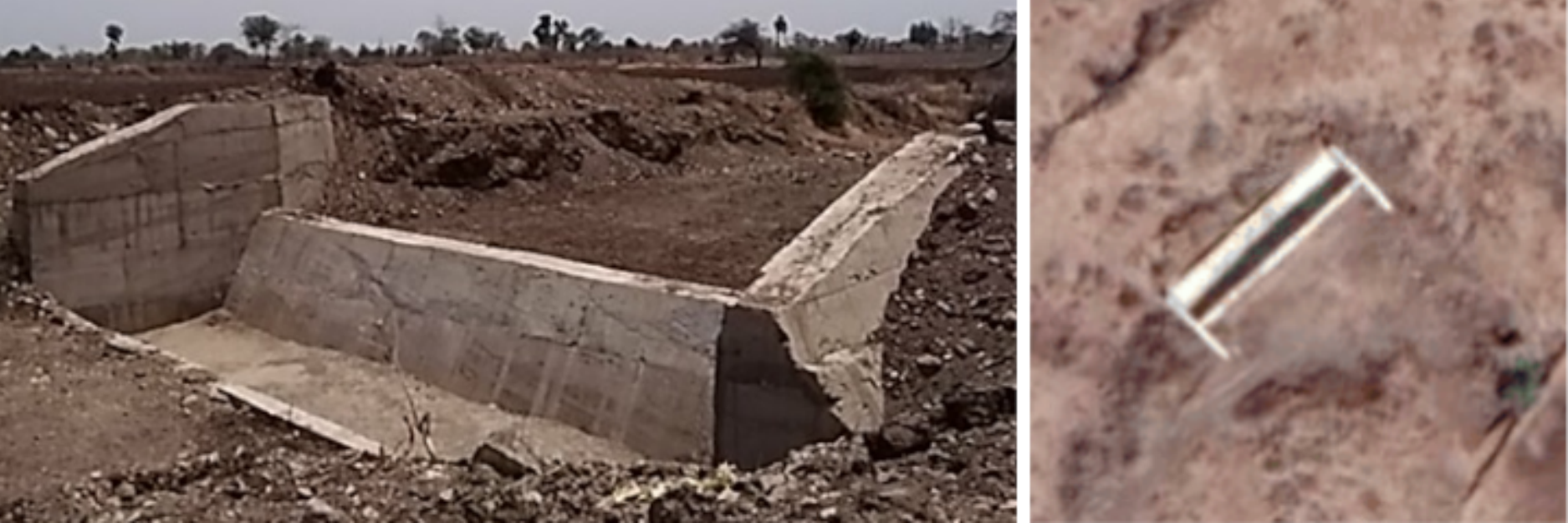}
     \caption{Photograph (left) and satellite image (right) of a wall based check-dam.}
     \label{fig:compare_wallbasedCD}                   
     \end{center}
\end{figure}
\begin{figure}[h]
     \begin{center}
     \includegraphics[width=0.98\linewidth]{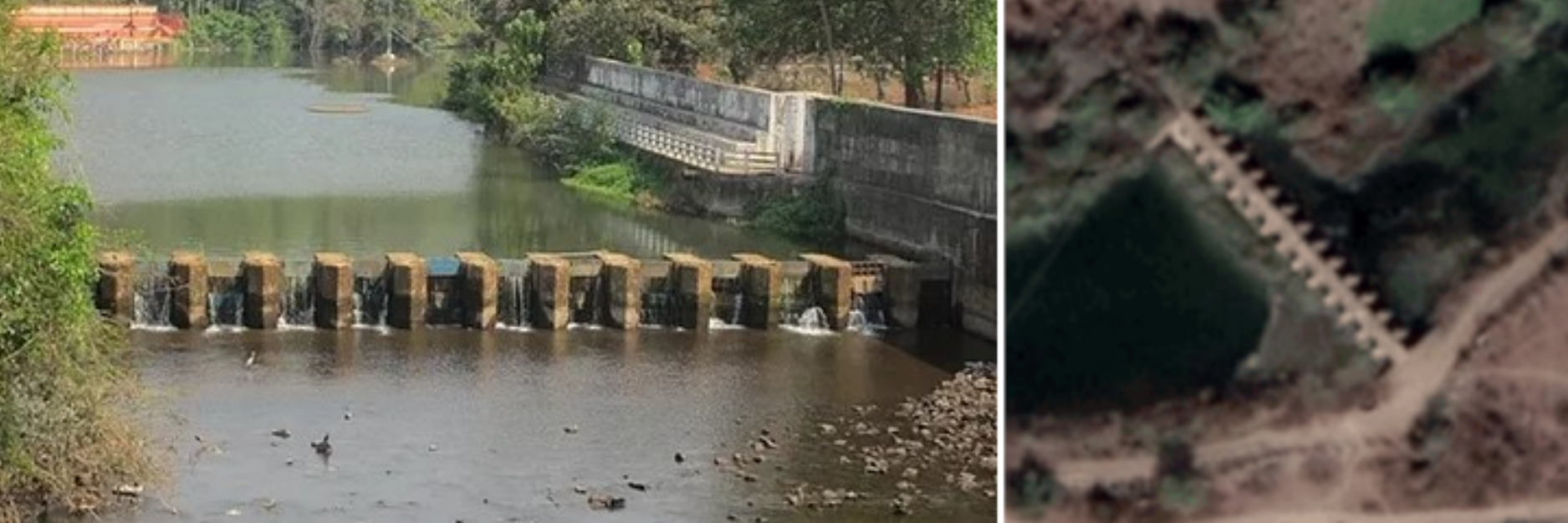}
     \caption{Photograph (left) and satellite image (right) of a gate based check-dam.}
     \label{fig:compare_gatebasedCD}                   
     \end{center}
\end{figure}

In general, images from satellites have small, densely clustered objects of interest with a small spatial extent, unlike that of COCO \cite{lin2014microsoft} or Imagenet \cite{russakovsky2014imagenet} with large and less clustered objects. For e.g. In satellite images, objects like cars will be a few tens of pixels in extent even at the highest resolution. In addition, the object images captured from above can have different orientations. i.e. objects classes like planes, vehicles, etc. can be oriented between 0 and 360 degrees, whereas object classes in COCO or Imagenet data are mostly vertical. 

Recent works~\cite{gistam20} have focused and attended to identifying man-made agricultural and irrigation structures used for farming. Check-dams (See Fig.\ref{fig:compare_wallbasedCD} and Fig.\ref{fig:compare_gatebasedCD}) are artificial irrigation structures built across water bodies to provide irrigation to nearby and surrounding places. They have a cultivable command area of up to 2000 hectares. Minor irrigation census is conducted for purposes of rural planning, formulating various policies and in reserving and allocating funds for agricultural development. However, maintaining a record of the minor irrigation structures makes irrigation census a time-consuming, expensive and laborious task, which is prone to errors. In the light of this, we introduce a new dataset of check-dams from satellite images for building an automated system for detection and mapping of check-dams to ease the process of census. Our work will complement existing object detection techniques on agricultural structures~\cite{gistam20} like farm ponds and wells. In this paper, we primarily focus on deep learning based object detection and instance segmentation methods and benchmark the latest methods in computer vision. Our contribution in this paper is two-fold :
\begin{enumerate} 
\item
\textbf{M}inor \textbf{I}rrigation \textbf{S}tructures \textbf{C}heck-\textbf{D}am Dataset: a public dataset annotated by domain experts using images from Google static map APIs~\cite{website:googleMaps} for instance segmentation and object detection tasks. 
\item A benchmark and assessment of the performance of various object detection and instance segmentation methods on the check-dam dataset.
\end{enumerate}
The paper is structured as follows: section~\ref{sec:Related Work} covers the related work with an overview of the object detection and instance segmentation methods, section~\ref{sec:dataset} covers the details of the proposed dataset, section~\ref{sec:implementation} covers the evaluation criteria, architecture and training details, section \ref{sec:exp} covers the experiments, results, observation and analysis, and finally conclusion in section \ref{sec:conclusion}.

\section{\uppercase{Related Work}}
\label{sec:Related Work}
Deep learning with convolutional layers has led to learning of high-level feature representations of images, followed by the trend of overlapping detectors and proposal generators leading to powerful object detectors. On the other hand, instance segmentation gives a finer inference for every pixel of the object in the input image, labelled using the segmentation labels. 

\subsection{Object Detection Methods}
In general, object detection based methods are either based on region proposals or regression. Object proposals refer to the candidate boxes that possibly contain objects and are used to detect objects of multiple aspect ratios while avoiding sliding window searches. In two stage models, the first stage generates object proposals, followed by the refinement of these bounding boxes in the subsequent stages. In single stage models, the whole detection pipeline is performed in a single step, formulated as dense classification, generally optimized by a focal loss and localization of the object as bounding box regression. These regression based methods do not produce candidate region proposals or use any feature re-sampling methods, which lead to more efficient models. In bounding box regression, the location of a predicted bounding box is refined based on the anchor box and has been integrated into the detector while training. Deep regression applies deep learning to directly regress the bounding box coordinates based on deep learning features, but it is prone to difficulties in the localization of small objects. In this section, we briefly mention some of the object detection methods and summarize the important changes that each method introduced. Many of these object detection methods also double as instance segmentation methods.

RCNN~\cite{girshick2013rich} extracts a set of object proposals by a selective search on a given image, and then the region proposals are re-scaled and fed to a backbone to extract features. These features are then used by SVM classifiers to predict and recognize the object categories within a region. Fast RCNN~\cite{girshick2015fast} simultaneously trains a detector and performs bounding box regression, leading to detection speeds over 200 times faster than the RCNN. However, most of its computational time and resources get expended in computing the fully connected layers. Later, faster region based CNN (Faster RCNN)~\cite{ren2015faster} introduces Region Proposal Method (RPN) and a detector that runs detection only on the network's top layer to further improve the inference speed.

YOLO~\cite{redmon2015look} introduces the single stage detection paradigm without using proposal detection and verification. Though YOLO is extremely fast, it lacks the localization accuracy in comparison to two-stage object detection methods as well as the performance on small objects. SSD~\cite{ssd} focuses on YOLO's issues by focusing on different layers of the network rather than just the topmost layer. YOLOv2~\cite{yolov2} improves the detection of small objects over YOLO by using larger input sizes. YOLOv3~\cite{redmon2018yolov3} improves on the various metrics over YOLOv2~\cite{yolov2} and improves on the real-time inference and execution time. RetinaNet~\cite{lin2017retina} solves the issue of foreground-background class imbalance during training of dense one-stage detectors, by reshaping the cross entropy loss to focus on hard examples and down-weighing the loss assigned to well classified examples. 

A few of the single-stage methods alleviate the anchor box imbalance between positive and negative anchor boxes by treating bounding boxes as keypoints pair. Cornernet~\cite{law2018cornernet}, a keypoint based one-stage detector eliminates the need for designing anchor boxes by using corner pooling to improve the corner localization. CascadeRPN~\cite{vu2019cascade_rpn} improves the quality of region-proposals by relying on single anchor per location and refines it in multiple stages, where each stage progressively defines positive samples by starting with anchor-free metric followed by anchor-based metrics. Free Anchor~\cite{zhang2019freeanchor} focuses on the idea of allowing objects to match the anchors flexibly and optimizes the detection customized likelihood.  Fully Convolutional One Stage (FCOS) object detector~\cite{tian2019fcos}, an anchor-free, proposal-free method approaches object detection in a per-pixel fashion, with only NMS post-processing step. Empirical Attention~\cite{zhu2019empirical} compares spatial attention elements from Transformer attention, dynamic convolution and deformable convolution in a generalized attention formulation. CentripetalNet~\cite{CentripetalNet} improves the corner point matching by using centripetal shift and by designing cross-star deformable convolution. DetectoRS~\cite{qiao2020detectors} combines Recursive Feature Pyramid (RFP) and Switchable Atrous Convolution (SAC) to achieve SoTA object detection performance. RFP incorporates feedback connections into the bottom-up backbone layers from the feature pyramid networks (FPN), while SAC uses the different atrous gates to convolve the features and uses switch functions to gather the results.

SparseRCNN~\cite{peize2020sparsercnn} reduces the number of hand-designed object candidates from thousands to a few hundred learnable proposals and predicts the final output directly without NMS post-processing. Dynamic RCNN~\cite{DynamicRCNN} proposes a dynamic design to alleviate the inconsistency problem between the fixed network and dynamic training procedures by adjusting the IoU threshold and the shape of regression loss function, based on the training statistics of the object proposals. GFL~\cite{li2020gfl} designs a joint representation of localization quality and classification to eliminate the inconsistency risk and depicts the flexible distribution in real data, by merging the quality estimation into class prediction vector. The resulting labels are continuous, which are optimized using a generalized focal loss. Deformable DETR~\cite{zhu2021deformabledetr} mitigates the issues of DETR by attending to key sampling points around a reference to overcome slow convergence and the limitations of feature spatial resolution.
\subsection{Instance Segmentation Methods}
In this section, we briefly mention some of the instance segmentation methods and summarize the important changes in each method. All these instance segmentation methods can also perform object detection under the same pipeline.

Cascade RCNN~\cite{Cai_2019_cascadercnn} uses a sequence of detectors that are trained sequentially using the output of a detector as the training set for the next detector. In cascaded detection, a coarse to fine technique is used, improving localization accuracy for small objects. Instaboost~\cite{fang2019instaboost}, a data augmentation technique, explores feasible locations where objects could be placed based on the similarity of local appearances and proposes a location probability map. Hybrid Task Cascade~\cite{chen2019htc} introduces a framework for joint multi-stage processing of both detection and segmentation pipelines by progressively learning discriminative features by integrating complementary features.

YOLACT~\cite{yolact-iccv2019} performs real-time instance segmentation by having two parallel pipelines, where one generates a set of prototype masks, while the other predicts the coefficients of masks per-instance. General ROI Extraction~\cite{rossi2020GRoI} introduces non-local building blocks and attention mechanisms to attend to multiple layers of FPN, to extract a coherent subset of features for integrating in two stage methods. Prime Sample Attention (PISA) in object detection~\cite{cao2019pisa} assesses how different samples from the dataset contribute to the overall performance in mean AP. Swin Transformer~\cite{liu2021Swin} presents a hierarchical architecture that uses shifted windows for computing representations by limiting self-attention computation to non-overlapping local windows, while also permitting cross-window connections.

\section{\uppercase{Proposed Dataset}}
\label{sec:dataset}
Here, we introduce and give the details of our dataset of satellite images on check-dams, an irrigation structure constructed for agricultural needs. Our dataset has images comparable in image dimensions to that of Imagenet and COCO, instead of commonly used single-view remote sensing image with dimensions in the range of tens of thousands of pixels. Our dataset falls in the category of small-scale optical satellite image dataset, whose ground truth instance level annotations are not easily available. Our dataset complements minor irrigation structures dataset~\cite{gistam20} for wells and farm ponds, by adding a new category in man-made irrigation structures. While most datasets are designed keeping only one task in mind, our dataset is designed with annotations for both object detection and instance segmentation tasks. 
To ensure geographical diversity in the dataset, we collected images from 36 districts of Maharashtra, India, based on the availability of ground-truth data. 

\noindent\textbf{Related Datasets:}
There are several datasets in remote sensing that use images from multiple sources like airplane, drone (aerial images), satellites (optical, multispectral, hyperspectral, etc.). Some of these datasets focus on objects like buildings, vehicles, cars, ships, etc., while some have classes like pools, playgrounds, vehicles of different types, etc. While some datasets have very high resolution images, some datasets have images ranging from 10~\cite{maggiori2017dataset} to a few tens of thousands. Datasets also vary in terms of the number of instances from 600~\cite{maggiori2017dataset} to around 1/5 of a million~\cite{diordataset}. A comparison of the related datasets in terms of the number of classes, number of instances, number of images and the image sizes are given in the Table \ref{tab:dataset}.

\begin{table}[h]
\centering
\caption{Optical satellite image datasets for object detection.}
\resizebox{0.46\textwidth}{!}{
\begin{tabular}{|c|c|c|c|c|}
\hline
\textbf{Dataset} & \textbf{\#Classes} & \textbf{\#Instances} & \textbf{\#Images} & \textbf{\begin{tabular}[c]{@{}c@{}}Image Size\\ (pixels)\end{tabular}} \\ \hline
\begin{tabular}[c]{@{}c@{}}RSOD\\~\cite{rsod} \end{tabular}
& 4 & 6,950 & 976 & $\sim$1000   \\ \hline
\begin{tabular}[c]{@{}c@{}}CARPPK\\~\cite{Hsieh_2017_ICCV}\end{tabular}
& 1 & 89,777 & 1448 & 1280 \\ \hline
\begin{tabular}[c]{@{}c@{}}ITCVD\\~\cite{itcvd}\end{tabular}
& 1 & 228 & 23,543 & 5616 \\ \hline
\begin{tabular}[c]{@{}c@{}}LEVIR\\~\cite{Chen2020} \end{tabular}
& 3 & 11,000 & 22,000 & 600-800 \\ \hline
\begin{tabular}[c]{@{}c@{}}SpaceNet MVOI\\~\cite{spacenet} \end{tabular} & 1 & 126,747 & 60,000 & 900 \\ \hline
\begin{tabular}[c]{@{}c@{}}DIOR\\ ~\cite{diordataset} \end{tabular}
& 20 & 90,228 & 23,463 & 800 \\ \hline
\begin{tabular}[c]{@{}c@{}}RarePlanes\\~\cite{RarePlanes_Paper} \end{tabular}
& 1 & 644,258 & 50,253 & 1080 \\ \hline
\begin{tabular}[c]{@{}c@{}}Well \\~\cite{gistam20} \end{tabular} & 1 & 1614 & 1,011 & 640 \\ \hline
\begin{tabular}[c]{@{}c@{}}Farmpond \\~\cite{gistam20} \end{tabular}     & 4 & 715 & 1,018 & 640 \\ \hline
\textbf{MIS Check-Dam}                                           & 2 & 1082 & 1,037 & 640 \\ \hline
\end{tabular}}
\label{tab:dataset}
\end{table}
\noindent \textbf{Scale}:
Contrasting the datasets~\cite{diordataset} that have large class imbalance, we have 562 instances of wall-based check-dam and 520 instances of gate-based check-dam.  While other datasets have relatively abundant and dense object instances that are easy to capture, our dataset has instances that are geographically far apart from each other and require annotations from the domain experts. In comparison to dense object datasets (e.g. cars in aerial images), our dataset has images consisting of sparsely located objects. The object instances in our dataset come from two categories. Our dataset has 1082 instances of check-dams across 1037 images on images of 640 x 640 pixel dimensions. Our dataset also has images with visually similar objects to check-dams like buildings, huts, etc. that makes the object detection and instance segmentation task difficult. In terms of spatial resolution, Google maps API uses zoom levels ranging from 1 to 22 that correspond to the image spatial resolution. However, most objects of interest are identifiable to humans at zoom levels of 17, 18 and 19 with resolutions of 1.262, 0.315 and 0.078 respectively. 
While a heavy traffic image can have up to hundreds of instances, only 8-10 farm ponds and wells are visible at zoom level 18, and upto four check-dams are visible at zoom level 18. \\
\noindent \textbf{Size Variations}:
Check-dams are subject to size variations due to their surrounding environments. Unlike other dataset classes, our dataset has instances with large intra-class diversity and high inter-class similarity. Since a check-dam is built across a river or water body, it is easy for a human with domain knowledge to identify the structure and distinguish between a check-dam or a bridge. However, the width of a water body varies along it's trajectory giving rise to large variations in check-dam sizes in comparison to objects like cars or planes in other datasets, which may have a standard shape and size.

\noindent \textbf{Image Variations:}
The satellite images collected from Google maps span seasons and collection times. The imaging conditions also vary due to the presence of air pollution, clouds, weather conditions and image sensor types, which leads to blurry images or image tinting. Moreover, there are wide variations in the surrounding vegetation even under a small geographical region. Unlike nadir (overhead) satellite imagery, off-nadir imagery captures different perspectives of the same object. An object captured from one angle may cast the shadow and appear differently based on the time of image capture. Also, the presence of trees and surrounding vegetation may occlude the check-dam. Moreover, a check-dam and its surrounding appearances vary depending upon the water level, seasonal rains and the irrigation purposes and uses.

\begin{table*}[h]
\centering
\caption{Comparison of the performance of various instance segmentation methods on checkdam dataset.}
\resizebox{0.98\textwidth}{!}{%
\begin{tabular}{|c|c|c|c|c|c|c|c|c|}
\hline
\textbf{Method} &
  \textbf{Model Name} &
  \textbf{Backbone} &
  \textbf{\begin{tabular}[c]{@{}c@{}}bbox mAP \\ (0.50:0.95)\end{tabular}} &
  \textbf{\begin{tabular}[c]{@{}c@{}}bbox mAP\\  (0.50)\end{tabular}} &
  \textbf{\begin{tabular}[c]{@{}c@{}}bbox mAP \\ (0.75)\end{tabular}} &
  \textbf{\begin{tabular}[c]{@{}c@{}}segm mAP \\ (0.50:0.95)\end{tabular}} &
  \textbf{\begin{tabular}[c]{@{}c@{}}segm mAP \\ (0.50)\end{tabular}} &
  \textbf{\begin{tabular}[c]{@{}c@{}}segm mAP\\  (0.75)\end{tabular}} \\ \hline
\multirow{2}{*}{Hybrid Task Cascade \cite{chen2019htc} } &
  HTC &
  R-50 &
  0.605 &
  0.940 &
  0.705 &
  \textbf{0.478} &
  0.918 &
  0.472 \\ \cline{2-9} 
 &
  HTC &
  R-101 &
  0.599 &
  0.936 &
  0.705 &
  0.474 &
  0.905 &
  0.446 \\ \hline
\multirow{2}{*}{YOLACT \cite{yolact-iccv2019}} &
  YOLACT &
  R-50 &
  0.539 &
  0.966 &
  0.542 &
  0.458 &
  \textbf{0.939} &
  0.409 \\ \cline{2-9} 
 &
  YOLACT &
  R-101 &
  0.545 &
  0.940 &
  0.574 &
  0.435 &
  0.908 &
  0.377 \\ \hline
\multirow{2}{*}{Instaboost \cite{fang2019instaboost}} &
  Cascade MRCNN &
  R-50 &
  0.613 &
  0.933 &
  \textbf{0.722} &
  0.477 &
  0.915 &
  0.451 \\ \cline{2-9} 
 &
  Cascade MRCNN &
  R-101 &
  0.607 &
  0.942 &
  0.707 &
  0.457 &
  0.898 &
  0.444 \\ \hline
\multirow{2}{*}{Cascade RCNN \cite{Cai_2019_cascadercnn} }  &
  \begin{tabular}[c]{@{}c@{}}Cascade MRCNN\end{tabular} &
  R-50 &
  0.601 &
  0.932 &
  0.692 &
  0.466 &
  0.902 &
  0.449 \\ \cline{2-9} 
 &
  \begin{tabular}[c]{@{}c@{}}Cascade MRCNN\end{tabular} &
  R-101 &
  0.567 &
  0.933 &
  0.638 &
  0.460 &
  0.893 &
  0.423 \\ \hline
General ROI Extraction \cite{rossi2020GRoI} &
  MRCNN &
  R-50 &
  0.587 &
  0.950 &
  0.697 &
  0.472 &
  0.908 &
  0.441 \\ \hline
Prime Sample Attention \cite{cao2019pisa} &
  MRCNN &
  R-50 &
  0.603 &
  0.941 &
  0.703 &
  0.477 &
  0.914 &
  \textbf{0.477} \\ \hline
\multirow{2}{*}{Swin Transformer \cite{liu2021Swin}} &
  MRCNN (FP-16) &
  \begin{tabular}[c]{@{}c@{}}Swin Transformer\end{tabular} &
  0.603 &
  0.971 &
  0.699 &
  0.465 &
  0.921 &
  0.451 \\ \cline{2-9} 
 &
  MRCNN &
  \begin{tabular}[c]{@{}c@{}}Swin Transformer\end{tabular} &
  \textbf{0.617} &
  \textbf{0.978} &
  0.697 &
  \textbf{0.478} &
  0.920 &
  0.464 \\ \hline
\end{tabular}%
}
\label{tab:segmentation-results}
\end{table*}

\section{\uppercase{Implementation Details}}
\label{sec:implementation}
\subsection{Evaluation Criteria}
The two main metrics used for object detection and instance segmentation tasks are PascalVOC metric~\cite{pascal-voc-2012} and COCO metric~\cite{lin2014microsoft}. These metrics rely on Intersection Over Union (IOU) given by the overlapping area between the predicted bounding box and the ground truth bounding box, divided by the area common between them.

PascalVOC metric takes the average precision at a fixed IoU of 0.5, whereas COCO metric measures the average over multiple IoU thresholds.
For multi-class detectors, final mAP (AP@[0.5:0.95]) is computed by taking the mean over 10 IoU levels (starting from 0.5 to 0.95 with a step size of 0.05) on each class. We base all our experiments on COCO metric, since COCO metric provides metrics at different IOU thresholds.
\subsection{Architecture and Training Details}
For training our dataset, we made a train-test split of 80-20, which consists of 450 wall-based check-dam instances and 418 gate-based check-dam instances in the training set and 110 wall-based check-dam instances and 103 gate-based check-dam instances in the test set. The details of architecture and backbone used in some of the methods are given in Table \ref{tab:model-architectures}. The specific details of the method, the model, and the backbone used for training are given in Table \ref{tab:segmentation-results} and Table \ref{tab: detection-results}. We used MMDetection~\cite{mmdetection} framework for implementing the various methods in our benchmark. We used ResNet 50, ResNet 101, ResNext 101, HourglassNet, DetectoRS-R50 pretrained backbones in object detection experiments and Resnet-50, Resnet-101, and Swin Transformer pretrained backbones in instance segmentation experiments. We used pretrained backbones primarily trained on Imagenet, COCO, and cityscapes for training different models.

\begin{figure*}[h]
 \begin{center}
 \includegraphics[width=0.98\linewidth]{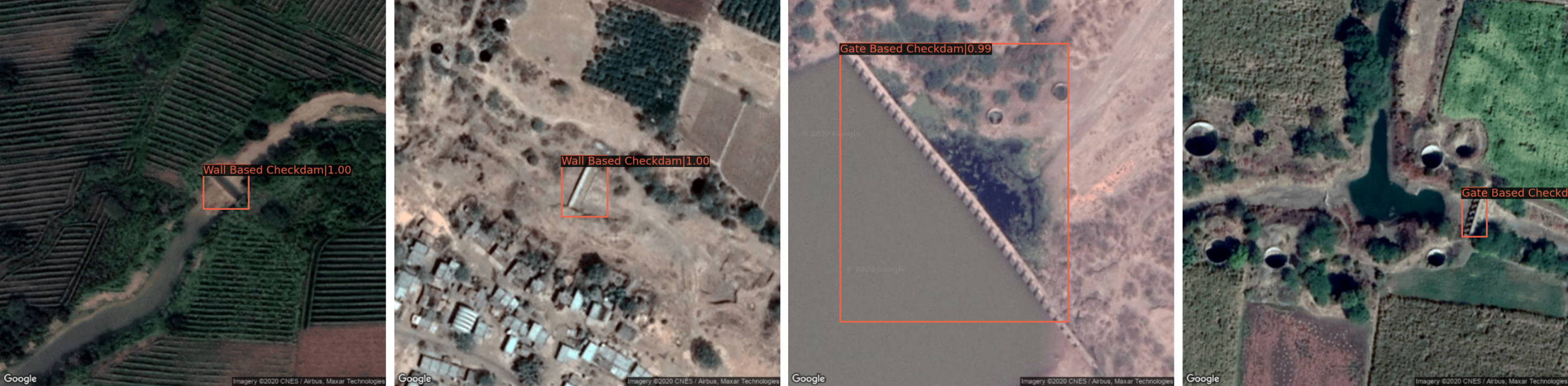}
 \caption{Object Detection results on the test data. (From left to right) a, b: Wall based check-dams detected among vegetation and dry surroundings; c,d: Detection on large size varieties of gate based check-dams.}
 \label{fig:detres1}                   
 \end{center}
\end{figure*}
\begin{figure*}[h]
\begin{center}
\includegraphics[width=0.985\linewidth]{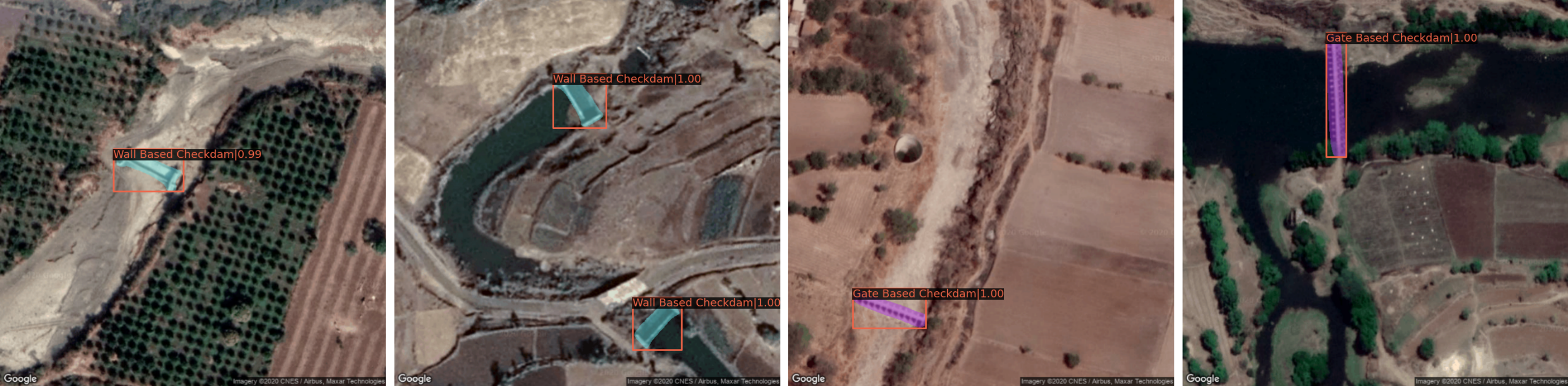}
\caption{Instance segmentation results on the test data. (From left to right) a, b: Wall based check-dams segmented in dry stream and wet stream surroundings; c, d: Gate based check-dams segmented in dry stream and wet stream.}
\label{fig:segres1}                   
\end{center}
\end{figure*}

\begin{table}[!ht]
\centering
\caption{Summary of the different architectures used in the comparison study.}
\resizebox{0.45\textwidth}{!}{%
\begin{tabular}{|c|c|}
\hline
\textbf{Method} & \textbf{Architecture Details} \\ \hline
YOLOv3 & Darknet \\ \hline
CornerNet & \begin{tabular}[c]{@{}c@{}}Stacked Hourglass, Corner pooling\end{tabular} \\ \hline
Centripetal Net & \begin{tabular}[c]{@{}c@{}}Stacked Hourglass, Corner pooling\end{tabular} \\ \hline
Empirical Attention & \begin{tabular}[c]{@{}c@{}}Deformable Convolution(DCN), \\ Spatial Attention(1111, 0010)\end{tabular}          \\ \hline
FCOS & \begin{tabular}[c]{@{}c@{}}Group Normalization(GN), DCN \end{tabular} \\ \hline
Generalized Focal Loss & \begin{tabular}[c]{@{}c@{}}Generalized Focal Loss, DCNv2\end{tabular} \\ \hline
General ROI Extraction & \begin{tabular}[c]{@{}c@{}}Non-local building block + Attention\end{tabular} \\ \hline
Prime Sample Attention & \begin{tabular}[c]{@{}c@{}}PISA, ROI Pool\end{tabular} \\ \hline
Hybrid Task Cascade & \begin{tabular}[c]{@{}c@{}}DCN, Multiscale Training\end{tabular} \\ \hline
DetectoRS & \begin{tabular}[c]{@{}c@{}}RFP, SAC \end{tabular} \\ \hline
\end{tabular}
}
\label{tab:model-architectures}
\end{table}

\section{\uppercase{Experiments and Results}}
\label{sec:exp}
\textbf{Localization Distillation:}
The network size and the number of parameters are crucial for network inference speeds. Network distillation compresses the knowledge from a teacher network to a smaller student network, by overcoming the limitation of distilling only localization information, for arbitrary teacher and student architectures. While network pruning reduces the network sizes by removing the unimportant weights, this cannot be applied directly to object detection methods, as it may result in sparse connectivity patterns in a CNN. We experiment to see whether knowledge distillation is possible with localization distillation~\cite{zheng2021LD}. We used three different settings based on GFL~\cite{li2020gfl}, training a teacher with a pretrained backbone of ResNet-101 and a student with different backbone settings of ResNet-18, ResNet-34 and ResNet-50. The details of the teacher-student backbone and the results of the knowledge distillation on object detection task are given in Table \ref{tab:localization-distillation}. We can observe from Table \ref{tab:localization-distillation} that the bbox mAP (0.50) of 0.966 is close to 0.967 of DetectoRS~\cite{qiao2020detectors} given in Table \ref{tab: detection-results}. 
Also, performing the localization distillation on the same model as the teacher can further improve the performance of the model. 
\subsection{Results}
The quantitative results of comparison of various object detection methods are given in Table \ref{tab: detection-results} and the results of comparison of various instance segmentation methods are given in Table \ref{tab:segmentation-results}. The results of training a Faster-RCNN model from scratch is given in the topmost row of Table \ref{tab: detection-results} while the remaining table rows compare the performance of methods trained using pre-trained backbone. We observe that the model trained from scratch has performed poorly even after training for more than a hundred epochs, while the metric scores have improved only gradually. We observe that using pre-trained backbone not only improves the convergence speed, but also helps to achieve the best possible performance from a method. We observe that DetectoRS has the highest values of 0.625 in mAP (0.50:0.95), 0.967 in mAP (0.5) and 0.749 in mAP (0.75) for the object detection task. We attribute the performance of DetectoRS to hybrid task cascade along with recursive feature pyramid and switchable atrous convolution. Though RFP and SAC can be independently applied, the best performance is observed when both are applied. For the instance segmentation task, both Swin transformer and HTC records the highest values of 0.478 in mAP (0.50:0.95), while YOLACT records the highest of 0.939 in mAP (0.5) and PISA records the highest of 0.477 in mAP (0.75).

The qualitative results of object detection by DetectoRS on our dataset are given in Fig. \ref{fig:detres1}, while the qualitative results of instance segmentation using Hybrid Task Cascade are given in Figure \ref{fig:segres1}. From Fig. \ref{fig:detres1}, we can observe that DetectoRS is able to detect bounding boxes accurately for both small and large check-dams as well as in both green and dry surroundings. From Fig. \ref{fig:segres1}, we can observe that HTC is able to perform instance segmentation precisely on check-dams in both dry and wet streams.
\begin{table*}[h]
\centering
\caption{Comparison of the performance of various object detection methods on checkdam dataset.}
\resizebox{0.98\textwidth}{!}{%
\begin{tabular}{|c|c|c|c|c|c|}
\hline
\textbf{Method} &
  \textbf{Model Name} &
  \textbf{Backbone} &
  \textbf{\begin{tabular}[c]{@{}c@{}}bbox mAP\\  (0.50:0.95)\end{tabular}} &
  \textbf{\begin{tabular}[c]{@{}c@{}}bbox mAP \\ (0.50)\end{tabular}} &
  \textbf{\begin{tabular}[c]{@{}c@{}}bbox mAP \\ (0.75)\end{tabular}} \\ \hline
Faster  RCNN-Scratch &
  Faster RCNN  &
  R-50-FPN &
  0.145 &
  0.381 &
  0.077 \\ \hline
\multirow{2}{*}{Faster RCNN \cite{ren2015faster}} &
  Faster  RCNN &
  R-50-FPN &
  0.597 &
  0.948 &
  0.699 \\ \cline{2-6} 
 &
  Faster RCNN &
  R-101-FPN &
  0.576 &
  0.94 &
  0.625 \\ \hline
\multirow{2}{*}{YOLOv3 \cite{redmon2018yolov3}} &
  YOLOv3 &
  \begin{tabular}[c]{@{}c@{}}Darknet53 (Scale-416)\end{tabular} &
  0.504 &
  0.923 &
  0.503 \\ \cline{2-6} 
 &
  YOLOv3 &
  \begin{tabular}[c]{@{}c@{}}Darknet53 (Scale-608)\end{tabular} &
  0.526 &
  0.939 &
  0.588 \\ \hline
\multirow{2}{*}{RetinaNet \cite{lin2017retina}} &
  RetinaNet &
  R-50 &
  0.580 &
  0.940 &
  0.670 \\ \cline{2-6} 
 &
  RetinaNet &
  R-101 &
  0.586 &
  0.946 &
  0.681 \\ \hline
Free Anchor \cite{zhang2019freeanchor} &
  RetinaNet &
  R-50 &
  0.574 &
  0.939 &
  0.654 \\ \hline
Generalized Focal Loss \cite{li2020gfl} &
  GFL &
  R-101 &
  0.519 &
  0.918 &
  0.519 \\ \hline
CascadeRPN \cite{vu2019cascade_rpn} &
  FasterRCNN &
  R-50 &
  0.583 &
  0.945 &
  0.610 \\ \hline
Dynamic RCNN  \cite{DynamicRCNN}&
  FasterRCNN &
  R-50 &
  0.557 &
  0.951 &
  0.592 \\ \hline
\multirow{2}{*}{Empirical Attention \cite{zhu2019empirical} } &
  FasterRCNN (0010) &
  R-50 &
  0.594 &
  0.953 &
  0.667 \\ \cline{2-6} 
 &
  FasterRCNN (1111) &
  R-50 &
  0.597 &
  0.945 &
  0.710 \\ \hline
\multirow{2}{*}{FCOS Object Detector \cite{tian2019fcos} } &
  FCOS &
  R-50 &
  0.262 &
  0.612 &
  0.161 \\ \cline{2-6} 
 &
  FCOS &
  R-101 &
  0.191 &
  0.481 &
  0.116 \\ \hline
\multirow{4}{*}{SparseRCNN \cite{peize2020sparsercnn} } &
  SparseRCNN (\#Prop: 100) &
  R-50 &
  0.573 &
  0.935 &
  0.596 \\ \cline{2-6} 
 &
  SparseRCNN (\#Prop: 300) &
  R-50 &
  0.580 &
  0.944 &
  0.647 \\ \cline{2-6} 
 &
  SparseRCNN (\#Prop: 100) &
  R-101 &
  0.577 &
  0.939 &
  0.647 \\ \cline{2-6} 
 &
  SparseRCNN (\#Prop: 300) &
  R-101 &
  0.554 &
  0.937 &
  0.609 \\ \hline
\multirow{3}{*}{Cascade RCNN \cite{Cai_2019_cascadercnn} } &
  CascadeRCNN &
  R-50 &
  0.601 &
  0.939 &
  0.701 \\ \cline{2-6} 
 &
  CascadeRCNN &
  R-101 &
  0.593 &
  0.941 &
  0.658 \\ \cline{2-6} 
 &
  CascadeRCNN &
  ResNeXt-101 &
  0.600 &
  0.943 &
  0.685 \\ \hline
\multirow{2}{*}{CornerNet \cite{law2018cornernet}} &
  CornerNet (BS: 8x6) &
  HourglassNet &
  0.570 &
  0.894 &
  0.672 \\ \cline{2-6} 
 &
  CornerNet (BS: 32x3) &
  HourglassNet &
  0.575 &
  0.904 &
  0.662 \\ \hline
Centripetal Net \cite{CentripetalNet} &
  CornerNet (Batch Size 16x6) &
  HourglassNet &
  0.598 &
  0.943 &
  0.686 \\ \hline
\multirow{3}{*}{DetectoRS \cite{qiao2020detectors}} &
  CascadeRCNN &
  DectoRS R-50 &
  0.615 &
  \textbf{0.967} &
  0.711 \\ \cline{2-6} 
 &
  RFP - HTC + R-50 &
  DectoRS R-50 &
  0.617 &
  0.953 &
  0.693 \\ \cline{2-6} 
 &
  SAC - HTC + R-50 &
  DectoRS R-50 &
  \textbf{0.625} &
  0.966 &
  \textbf{0.749} \\ \hline
Deformable DETR \cite{zhu2021deformabledetr} &
  Two-stage Deformable DETR &
  R-50 &
  0.540 &
  0.930 &
  0.584 \\ \hline
\end{tabular}%
}
\label{tab: detection-results}
\end{table*}

\subsection{Observations and Analysis}
Recent works have integrated different techniques for improving the performance of object detection and instance segmentation tasks. 
Any improvement that reduces the number and size of the various network parameters without affecting the other aspects of detection or segmentation can speedup the inference stage.
In sliding window based detectors, the overlap between adjacent windows can be reduced by feature map shared computation of the whole image only once before sliding windows. 
Similarly, grouping the feature channels into independent groups can also reduce the parameter count. Another way to reduce the complexity of a layer and filters is to approximate it with fewer filters and a non-linear activation. Factorizing convolutions is an efficient way to replace a very large filter with smaller filter sizes, which can share the same receptive fields while being efficient. 
The local context can help improve the object detection by referring to the visual area that surrounds the objects of interest. Similarly the global context can help integrate the information of the different scene elements by having larger receptive fields or a global pooling of the CNN features, to improve the object detection performance.

\begin{table}[!h]
\centering
\caption{Comparing localization distillation performance using different neural network architectures between the (\textbf{T})eacher and the (\textbf{S})tudent.}
\resizebox{0.45\textwidth}{!}{%
\begin{tabular}{|c|c|c|c|}
\hline
\textbf{Backbone} & \textbf{\begin{tabular}[c]{@{}c@{}}bbox mAP \\ (0.50:0.95)\end{tabular}} & \textbf{\begin{tabular}[c]{@{}c@{}}bbox mAP\\  (0.50)\end{tabular}} & \textbf{\begin{tabular}[c]{@{}c@{}}bbox mAP \\ (0.75)\end{tabular}} \\ \hline
\begin{tabular}[c]{@{}c@{}}R-101 (T), R-18 (S)\end{tabular} & 0.532 & 0.934 & 0.572 \\ \hline
\begin{tabular}[c]{@{}c@{}}R-101 (T), R-34 (S)\end{tabular} & \textbf{0.592} & \textbf{0.966} & \textbf{0.670} \\ \hline
\begin{tabular}[c]{@{}c@{}}R-101 (T), R-50 (S)\end{tabular} & 0.566 & 0.957 & 0.654 \\ \hline
\end{tabular}
}
\label{tab:localization-distillation}
\end{table}

\section{\uppercase{Conclusions}}
\label{sec:conclusion}
We introduced MIS Check-Dam, a dataset for check-dams that belongs to the class of minor irrigation structures for agricultural use. We benchmark our dataset on some of the most recent object detection methods and instance segmentation methods. We also assess the importance of various components in the pipeline after evaluating these methods on our novel dataset. Future works can try domain adaptation methods on our dataset and other related datasets that have related classes like dams. 
\bibliographystyle{apalike}
{\small
\bibliography{bibliography}}

\end{document}